\documentclass[letterpaper, 10 pt, conference]{ieeeconf}  

\IEEEoverridecommandlockouts                              

\usepackage{amsmath}    
\usepackage{amssymb}    

\usepackage{mathtools}
\usepackage{mathrsfs}

\usepackage{gensymb,soul}
\usepackage{graphics} 
\usepackage{subfigure}
\usepackage{booktabs}
\usepackage{epsfig} 
\usepackage{cite}
\usepackage{mathptmx} 
\usepackage{amsfonts,amssymb, amsmath}  
\usepackage[colorlinks,linkcolor=blue]{hyperref}
\usepackage{tikz}

\usepackage{xcolor}
\usepackage{ifthen}
\usepackage{makecell}
\usepackage{algorithmicx}
\usepackage{algpseudocode}
\usepackage{threeparttable}
\usepackage[linesnumbered,ruled]{algorithm2e}

\DeclareMathAlphabet{\mathcal}{OMS}{cmsy}{m}{n}
\DeclareSymbolFont{largesymbols}{OMX}{cmex}{m}{n}

\usepackage{multirow}

\makeatletter
\let\NAT@parse\undefined
\makeatother

\title{\LARGE \bf
Embracing Bulky Objects with Humanoid Robots: Whole-Body Manipulation with Reinforcement Learning}
\author{Chunxin Zheng, Kai Chen, Zhihai Bi, Yulin Li, Liang Pan, Jinni Zhou, \\ Haoang Li, and Jun Ma, \textit{Senior Member, IEEE} 
 \thanks{Chunxin Zheng, Kai Chen, Zhihai Bi, Yulin Li, Jinni Zhou, Haoang Li, and Jun Ma are with The Hong Kong University of Science and Technology (Guangzhou), China (e-mail: jun.ma@ust.hk).}
\thanks{Liang Pan is with The University of Hong Kong, China (e-mail: liangpan1005@connect.hku.hk).}
}%

\begin{document}

\maketitle
\thispagestyle{empty}
\pagestyle{empty}

\begin{abstract}
Whole-body manipulation (WBM) for humanoid robots presents a promising approach for executing embracing tasks involving bulky objects, where traditional grasping relying on end-effectors only remains limited in such scenarios due to inherent stability and payload constraints. 
This paper introduces a reinforcement learning framework that integrates a pre-trained human motion prior with a neural signed distance field (NSDF) representation to achieve robust whole-body embracing. 
Our method leverages a teacher-student architecture to distill large-scale human motion data, generating kinematically natural and physically feasible whole-body motion patterns. This facilitates coordinated control across the arms and torso, enabling stable multi-contact interactions that enhance the robustness in manipulation and also the load capacity. 
The embedded NSDF further provides accurate and continuous geometric perception, improving contact awareness throughout long-horizon tasks. 
We thoroughly evaluate the approach through comprehensive simulations and real-world experiments. The results demonstrate improved adaptability to diverse shapes and sizes of objects and also successful sim-to-real transfer. 
These indicate that the proposed framework offers an effective and practical solution for multi-contact and long-horizon WBM tasks of humanoid robots.
\end{abstract}

\section{Introduction}

Recent advances in humanoid robots have facilitated their deployment in diverse domains, such as industrial applications and home services, where they are increasingly required to perform whole-body manipulation (WBM) tasks~\cite{9424385,8720243,fu2024humanplus}. In these applications, humanoid robots are unavoidably required to grasp and transport bulky objects. 
Unlike the manipulation of relatively small objects with an end-effector, humanoid robots can leverage the whole body (including arms and torso) to envelop and embrace objects in a human-like manner, which could distribute contact forces across the body to stably manipulate bulky objects. 
However, there are several key considerations for successful completion of such challenging tasks. First, it requires an advanced perception system that can accurately model the spatial relationships between the robot and objects in its environment. Also, it necessitates a highly effective method that empowers humanoid robots to acquire anthropomorphic behaviors through a whole-body control policy.

Essentially, WBM has witnessed significant success, which shows noteworthy performance through multi-contact strategies \cite{10611674}. 
It leverages the entire body and even environmental supports to execute more forceful, stable, and dexterous manipulation of bulky objects \cite{florek2014humanoid}. This makes it particularly effective for tasks that surpass the force or kinematic constraints of traditional grasping strategies with the end effector.
Existing WBM approaches typically rely on the simplified robot model and kinematic relationships to estimate the spatial positions of the robot's links (usually simplified as multiple line segments) \cite{liu2025opt2skillimitatingdynamicallyfeasiblewholebody}. 
However, such methods are unable to accurately model or perceive the precise geometry of the surfaces of the links, leading to significant errors in position estimation during contact-rich manipulation tasks. As a result, these inaccuracies could ultimately result in task failures.

 \begin{figure}[t]	
	\centering
	\includegraphics[width=1\linewidth]{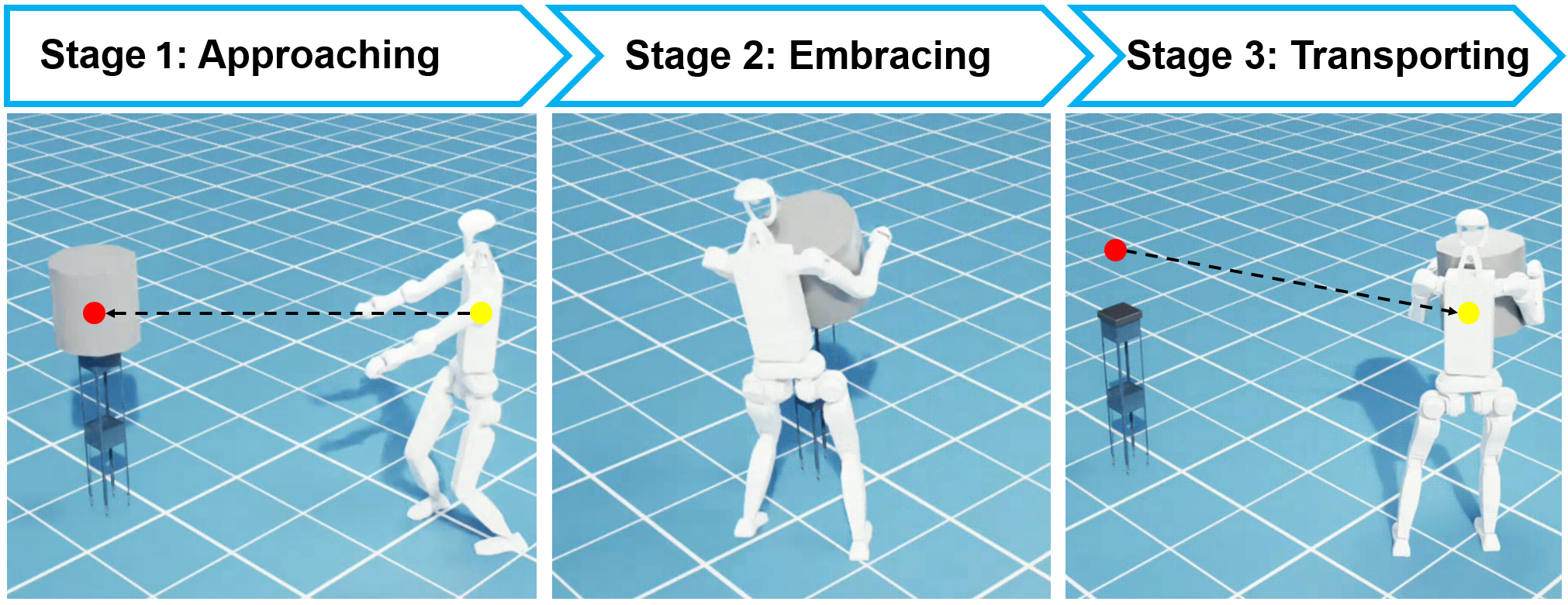}
	\caption
	{
    Whole-body manipulation task for a humanoid robot.
    The task mainly consists of three stages: (i) Approaching: the robot moves toward the object; (ii) Embracing: the robot establishes a whole-body enveloping grasp around the object; (iii) Transporting: the robot carries the object to a target location. The red dot denotes the object’s geometric center, and the yellow dot indicates the robot’s center of mass.}
    \label{fig:top}
\end{figure}

In general, control methods for WBM can be categorized into model-based \cite{7139995,9762117} and learning-based approaches \cite{8794160,zhang2023planguidedreinforcementlearningwholebody,doi:10.1126/scirobotics.ads6790}.
Model-based WBM methods need to specify the contact points and determine the contact sequence, but the computation is highly expensive due to the high degrees of freedom of humanoid robots. 
Learning-based methods, which generally depend on carefully engineered rewards, struggle to produce robust and coherent behaviors in these contact-intensive, dynamically unstable scenarios. In the existing literature, the developments on WBM for humanoid robots are still severely limited.
Especially for tasks like carrying bulky objects with WBM, it demands a sequence of coordinated actions as illustrated in Fig. \ref{fig:top}, which include approaching the object, embracing the object (by establishing multi-contact grasps and lifting with whole-body stabilization), and transporting objects to a designated location. 
In essence, acquiring the anthropomorphi skill of manipulating bulky objects with the whole body is extremely challenging due to: 1) the \textit{inherent high complexity} of humanoid robot systems with their numerous degrees of freedom, 2) the \textit{difficulty in constructing stable contact} through precise perception and control, and 3) the \textit{long-horizon nature} of manipulation tasks that require sustained coordination and planning.

To overcome these limitations, we propose a reinforcement learning (RL) framework that integrates a pre-trained human motion prior with a neural signed distance field (NSDF) representation, tailored to the whole-body embracing manipulation task of humanoid robots. 
The NSDF provides precise and efficient self-modeling capability that enhances the robot's spatial self-awareness and robustness of interactions. The human motion prior offers a biologically plausible kinematic distribution that stabilizes policy training and facilitates convergence in complex manipulation tasks.
Unlike conventional grasping with an end-effector only, our method dynamically coordinates the arms and torso of the humanoid robot to embrace bulky objects. In this sense, it overcomes the limitations of end-effector-only manipulation and significantly enhances payload capacity.
Crucially, the system maintains stability through distributed multi-contact interactions, ensuring balance while executing whole-body motions.
\begin{itemize}
    \item We propose an RL framework for humanoid robots to manipulate bulky objects using their entire bodies, including arms and torso. To the best knowledge of the authors, this is the first RL framework that enables the humanoid robots to actively engage with bulky objects through whole-body manipulation.

    \item   We introduce the motion prior into the embracing policy training pipeline, which speeds up the convergence in policy training toward such multi-contact and long-horizon tasks. Also, it empowers the humanoid robots to acquire anthropomorphic skills through the proposed framework.

    \item  We construct an NSDF representation of the humanoid robot that precisely perceives the robot-object interactions. It is further utilized to design both the observation space and reward function that guides the upper body to maintain sustained contact with the object, which significantly enhances the robustness of manipulation during the task.
    
    \item
    We thoroughly validate the proposed approach through simulations and real-world experiments on a humanoid robot, and the results attained demonstrate noteworthy performance in complex whole-body embracing tasks.
\end{itemize}

 \begin{figure*}[tp]	
	\centering
	\includegraphics[width=1\linewidth]{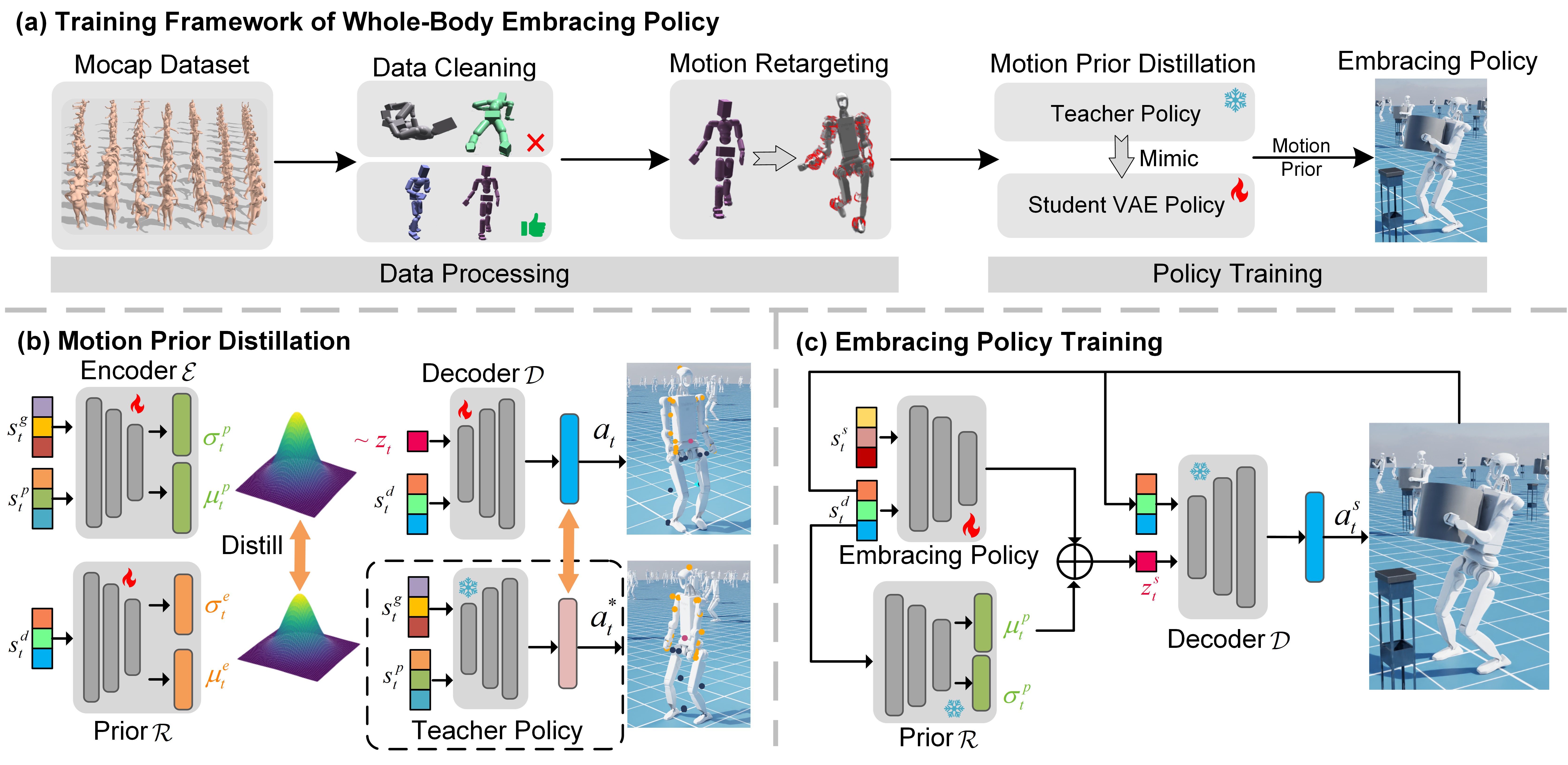}
	\caption{System architecture for training of embracing task with whole-body manipulation. (a) \textbf{Training Framework}:  The framework consists of two components, data processing and policy training. In data processing, we start from the AMASS mocap dataset, use MaskedMimic to discard sequences that violate kinematic constraints (e.g., self-collision and foot slip), and retarget the remaining motions to the robot morphology to collect robot-reference trajectories. In the policy training process, we first distill a human motion prior from the collected trajectories with a teacher policy and a student VAE policy, and then use the learned motion prior to guide the training of the high-level embracing policy. (b) \textbf{Motion Prior Distillation}: A pretrained teacher policy guides robotic motion imitation by providing reference actions. The student policy, structured as a VAE-based model, learns to reconstruct the teacher's action distribution through imitation learning. This process constructs a learnable humanoid motion prior $\mathcal{R}$, which facilitates efficient training for downstream tasks. (c) \textbf{Embracing Policy Training}: Based on the learned humanoid motion prior $\mathcal{R}$, we train the whole-body manipulation policy for the embracing task.}

    \label{fig:2}
\end{figure*}

\section{Related Work}

\subsection{Whole-Body Manipulation}
Considering the manipulation methods using only end-effectors as the interactive part, such as dexterous grippers and end sticks, WBM introduces the ability to manipulate objects using all possible effective parts of the robot. This approach represents the ultimate form of anthropomorphic manipulation by coordinating the whole body of a robot \cite{gu2025humanoid}. Despite this potential, current research largely focuses on single or bimanual manipulation platforms. Early works, such as \cite{932954}, employ model-based techniques to achieve whole-arm grasping on single-arm systems. Subsequently, research utilizing fixed-base humanoid platforms begin incorporating contact models involving dual arms, objects, and the torso to enable whole-arm manipulation \cite{florek2014humanoid}. However, the inherent complexity of contact-rich scenarios poses significant challenges for model-based methods. Consequently, recent approaches have turned to learning-based strategies. Notably, an RL method is presented in \cite{8794160} to achieve WBM on a bimanual platform. 
Also, some researchers have sought to enhance contact perception in WBM by equipping robot links with soft pneumatic sensors \cite{9762117}. These tactile enhancements provide richer contact feedback and facilitate more effective whole-body interaction. In the most recent state-of-the-art approaches, the integration of RL with whole-body tactile sensing has enabled more compliant and human-like motion generation for WBM on humanoid robots, but it is currently limited to upper-body platforms \cite{doi:10.1126/scirobotics.ads6790}.

\subsection{Reinforcement Learning for Humanoid Robot Control}

Significant progress has been made in humanoid robot control using RL, with extensive prior research dedicated to task-driven controllers for specialized behaviors, such as agile running \cite{zhuang2024humanoid} and locomotion across challenging terrains \cite{xie2025humanoid}. While these methods achieve high performance in their respective tasks, they often struggle to generalize to complex, long-horizon scenarios. Moreover, the bionicity of actions cannot be well guaranteed, or additional gait rewards need to be added during the training phase to guide the strategy. To address this limitation, an emerging trend leverages behavior cloning (BC) from physics-based animation \cite{peng2018deepmimic,10377518}, where control policies are trained on large-scale human motion datasets (e.g., AMASS). This approach enables robots to imitate natural, human-like movements while maintaining physical feasibility. Although BC policy has great performance on whole-body tracking, it typically requires an additional action generator \cite{ji2024exbody2,fu2024humanplus} or policy distillation \cite{he2024omnih2o} to adapt to specific tasks (e.g., locomotion, manipulation, or teleoperation). Moreover, they remain limited in interactive tasks due to the absence of critical environmental information in human motion datasets, such as contact forces, terrain geometry, and dynamic interactions, which are essential for robust real-world deployment.

Recent advancements in humanoid robotics have focused on developing universal motion controllers that can generalize across various tasks. These approaches often utilize pre-trained BC policies to construct a comprehensive action space. For example, Luo et al. \cite{luo2024universal} propose the use of human motion priors, distilled from BC policies, to create universal controllers. These controllers provide a robust foundation for further task-specific fine-tuning, facilitating the efficient adaptation of humanoid robots to diverse and complex tasks.
In \cite{langwbcrss2025}, it combines text description and pre-trained BC policy to build a text-motion mapping action space. Also, it has other methods by distilling different robotics skill policies to build action space servicing for universal humanoid robot controller \cite{zhang2025unleashinghumanoidreachingpotential}.


\section{Methodology}

In this work, we propose an RL framework for humanoid robots to perform bulky-object embracing tasks, and the complete system architecture is illustrated in Fig. \ref{fig:2}. In the following, we first present how to convert the AMASS mocap dataset to the kinematically feasible references of the robot (Section \ref{Data Processing}). Then, with these references, we detail the learning process of the humanoid motion prior through a teacher-student structure (Section \ref{sec:Motion Prior Distillation}), which provides a compact latent action space for downstream tasks. Next, in Section \ref{sec:Embracing}, we introduce the embracing policy training approach based on the learned motion prior. Pertinent notations used in this work are listed in Table \ref{tab:state_definition} for convenience.

\subsection{Data Processing}\label{Data Processing}

To enable humanoid robots to acquire fundamental movement skills from human demonstrations, the motion learning pipeline typically begins with the evaluation of raw motion capture data, represented using the SMPL human model. For example, MaskedMimic \cite{tessler2024maskedmimic} processes and filters these input motions to identify kinematically feasible trajectories. The process leads to a cleaned human motion dataset, denoted as $\mathcal{D}_h$, which contains only physically plausible movement patterns. We then adopt the retargeting framework from H2O \cite{he2024learning} to transfer these motions onto our target humanoid platform, explicitly accounting for differences in body proportions and joint ranges of motion. This produces the robot motion dataset, represented by $\mathcal{D}_r$, which provides validated trajectories to serve as training targets for the mimic teacher policy.

\begin{table}[t]
\centering
\caption{NOMENCLATURE}
\label{tab:state_definition}
\begin{tabular}{ccc}
\toprule
\textbf{Symbol} & \textbf{Dimension} & \textbf{Description} \\
\midrule
$\mathbf{a}_t$ & $\mathbb{R}^{27}$ & Target joint position \\
$\mathbf{v}_t$ & $\mathbb{R}^3$ & Root linear velocity \\
$\mathbf{\omega}_t$ & $\mathbb{R}^3$ & Root angular velocity \\
$\mathbf{g}_t$ & $\mathbb{R}^3$ & Projected gravity vector \\
$\mathbf{q}_t$ & $\mathbb{R}^{27}$ & Joint position \\
$\dot{\mathbf{q}}_t$ & $\mathbb{R}^{27}$ & Joint velocity \\
$\mathbf{a}_{t-1}$ & $\mathbb{R}^{27}$ & Previous action \\
$\hat{\mathbf{p}}_{t+1}$ & $\mathbb{R}^{3\times 27}$ & Reference body position \\
$\hat{\mathbf{p}}_{t+1} - \mathbf{p}_t$ & $\mathbb{R}^{3\times 27}$ & Positional offset \\
\bottomrule
\end{tabular}
\end{table}

\subsection{Motion Prior Distillation}\label{sec:Motion Prior Distillation}
The goal of this module is to distill the motion priors based on the robot motion dataset, which retain human-like dexterity while ensuring physical feasibility for humanoid robots. Following well-established techniques \cite{he2024omnih2o,luo2024universal}, we implement a teacher-student learning scheme. First, the retargeted robot motion dataset is utilized to train a teacher policy capable of precise trajectory tracking. Then, this teacher policy is distilled through a VAE-based student policy to construct the motion prior. The learned motion prior captures the essential characteristics of natural human movement, while remaining strictly aligned with the robot’s physical constraints.

\subsubsection{Teacher Policy}

Building upon the retargeted robot motion dataset, we develop a motion-tracking teacher policy using the PHC framework \cite{10377518}. The policy $\pi_{\text{teacher}}$ maps the current robot state and reference motion to actions:
\begin{align}
\mathbf{a}_t^{*} = \pi_{\text{teacher}}(\mathbf{s}^p_t, \mathbf{s}^g_t),
\end{align}
where $\mathbf{a}_t^{*}$ is the action generated by teacher policy, $\mathbf{s}^p_t = \{\mathbf{v}_t, \omega_t, \mathbf{g}_t, \mathbf{q}_t, \dot{\mathbf{q}}_t, \mathbf{a}_{t-1}^{*}\} \in \mathbb{R}^{90}$, which encodes the proprioceptive state through root linear velocity $\mathbf{v}_t$, angular velocity $\mathbf{\omega}_t$, projected gravity vector $\mathbf{g}_t$, joint positions $\mathbf{q}_t$, joint velocities $\dot{\mathbf{q}}_t$, and previous action $\mathbf{a}_{t-1}^{*}$. The goal state is defined as  $\mathbf{s}^{g}_t = \{\mathbf{\hat{p}}_{t+1}-\mathbf{p}_t, \mathbf{\hat{p}}_{t+1}\} \in \mathbb{R}^{2 \times 3 \times 27}$, which incorporates reference motion trajectories from $\mathcal{D}_r$, specified through both positional offsets ($\hat{\mathbf{p}}_{t+1} - \mathbf{p}_t$) and target configurations $\hat{\mathbf{p}}_{t+1}$ for each rigid body. The complete specifications of all variables, including their respective dimensions and detailed descriptions, are summarized in Table \ref{tab:state_definition}. The policy is implemented as a multi-layer perceptron (MLP) 
and optimized using Proximal Policy Optimization (PPO) \cite{schulman2017proximalpolicyoptimizationalgorithms}.

\subsubsection{Humanoid Motion Prior Distillation}

To enable efficient and transferable control of humanoid motion, we aim to distill the complex action space learned by the teacher policy into a compact latent representation that naturally induces human-like behavior. To this end, we adopt a VAE-based distillation approach, following PULSE~\cite{luo2024universal}, which encodes diverse motion behaviors into a latent space that can be effectively leveraged for downstream tasks.

Specifically, we construct a variational encoder-decoder framework to represent the humanoid motion space. The encoder $\mathcal{E}(\mathbf{z}_t \mid \mathbf{s}^p_t, \mathbf{s}^\text{g}_t)$ infers a distribution over the latent motion variable $\mathbf{z}_t$ given the current proprioceptive state $\mathbf{s}^p_t$ and the goal state $\mathbf{s}^\text{g}_t$. The decoder $\mathcal{D}(\mathbf{z}_t \mid \mathbf{s}^d_t, \mathbf{z}_t)$ maps the latent code back to the action space. To address sim-to-real transfer, we design $\mathbf{s}^d_t = \{ \omega_t, \mathbf{g}_t, \mathbf{q}_t, \dot{\mathbf{q}}_t, \mathbf{a}_{t-1}\} \in \mathbb{R}^{87}$, which excludes the root linear velocity component $\mathbf{v}_t$ from $\mathbf{s}^{p}_t$, making the representation feasible in real-world scenarios. Here, we use $\mathbf{a}_t$ to represent the action reconstructed by the student policy, and $\mathbf{a}_{t-1}$ to represent the previous student action at the last time. Furthermore, a learnable prior network $\mathcal{R}(\mathbf{z}_t \mid \mathbf{s}^{d}_t)$ is used to approximate the latent code distribution produced by the encoder. Formally, we have the following:
\begin{align}
\mathcal{E}(\mathbf{z}_t \mid \mathbf{s}^p_t, \mathbf{s}^{g}_t) &= \mathcal{N}(\mathbf{z}_t \mid \mu_t^e, \sigma_t^e), \\
\mathcal{R}(\mathbf{z}_t \mid \mathbf{s}^{d}_t) &= \mathcal{N}(\mathbf{z}_t \mid \mu_t^p, \sigma_t^p), \\
\mathcal{D}(\mathbf{a}_t \mid \mathbf{s}^{d}_t, \mathbf{z}_t) &= \mathcal{N}(\mathbf{a}_t \mid \mu_t^d, \hat{\sigma}_t^d),
\end{align}
where $\mathcal{D}$ outputs a Gaussian distribution over actions with a fixed diagonal covariance.

The overall training objective for the encoder, decoder, and prior is:
\begin{align}
\mathcal{L}_{\mathrm{all}} = \mathcal{L}_{\mathrm{action}} + \alpha \mathcal{L}_{\mathrm{regu}} + \beta \mathcal{L}_{\mathrm{KL}},
\end{align}
where $\mathcal{L}_{\mathrm{action}} = \|\mathbf{a}_t^{*} - \mathbf{a}_t\|_2^2$ is the action reconstruction loss. The regularization term, $\mathcal{L}_{\mathrm{regu}} = \|\mathbf{\mu}_t^{e} - \mathbf{\mu}_{t-1}^{e}\|_2^2$, penalizes abrupt changes in the latent trajectory by enforcing temporal consistency between consecutive latent means, where $\mathbf{\mu}_t^{e}$ and $\mathbf{\mu}_{t-1}^{e}$ are the mean of the encoder’s output distribution at time $t$ and $t-1$, respectively. The KL divergence term, $\mathcal{L}_{\mathrm{KL}}$, aligns the encoder’s latent distribution with the learned prior. The coefficients $\alpha$ and $\beta$ coordinate the relative importance of the smoothness and KL regularization terms, respectively.

Once the distillation is complete, we freeze the parameters of the decoder $\mathcal{D}$ and the prior $\mathcal{R}$, thus defining a new low-dimensional action space for downstream task learning.

\subsection{Embracing Policy Training}\label{sec:Embracing}

In this subsection, we formulate the task policy $\pi_\text{task}$ for the WBM task, especially for bulky objects. This task can be generally divided into three stages: approaching the object, embracing the object, and transporting the object to the target location.  To address this long-horizon manipulation task with bulky objects, we develop a tailored PPO algorithm incorporating three key components: stage randomized initialization, specific rewards, and humanoid motion prior distribution.

\subsubsection{Observations and Actions}

\begin{figure}[tp]	
	\centering
	\includegraphics[width=0.85\linewidth]{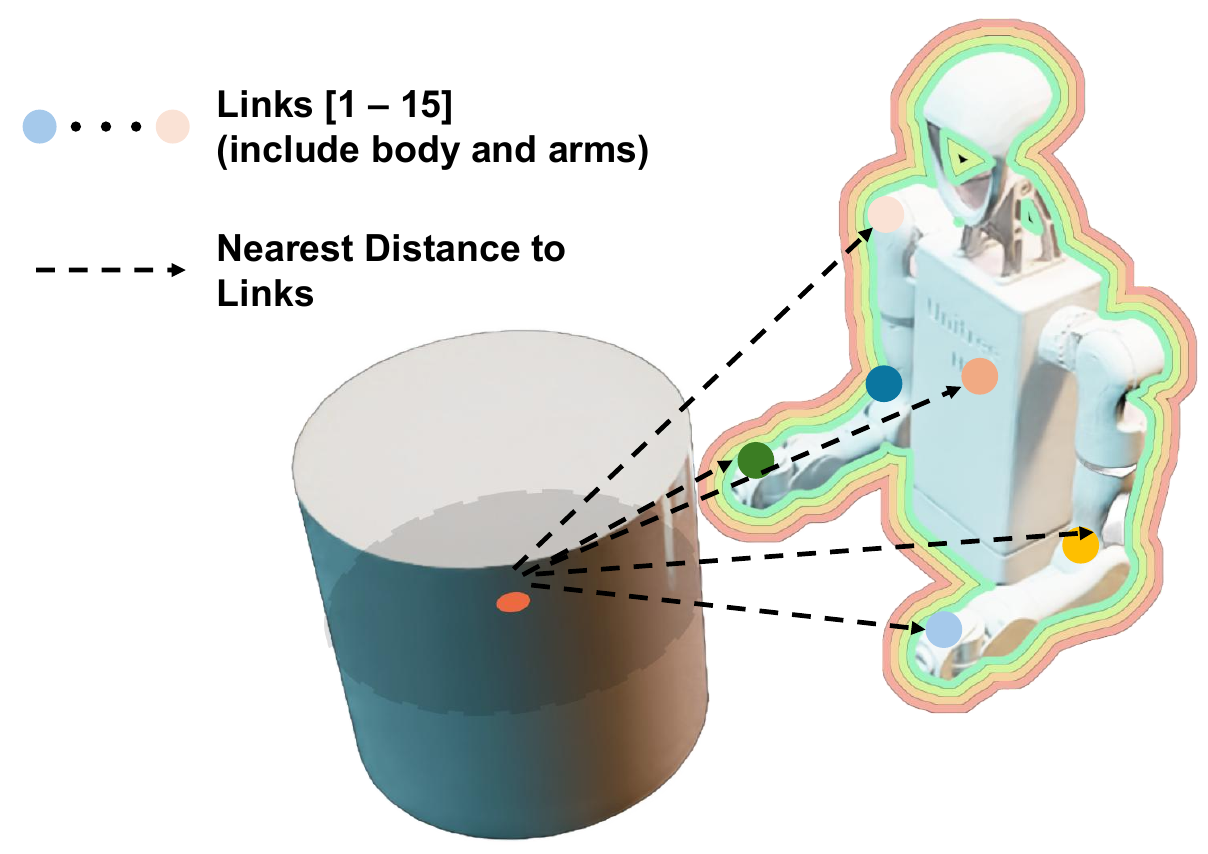}
	\caption{Illustration of NSDF between the target object and the robot's upper body. The dots distributed along each link of the robot represent the closest point from the corresponding joint to the object's center of mass. The dashed lines indicate the direction of the NSDF between the object's certer of mass and the robot’s links.}
    \label{fig:4s}
\end{figure}

The task observation at time step \( t \) is defined as \( \mathbf{s}_t^{\text{task}} = \{ \mathbf{s}_t^d,\, \mathbf{s}_t^s \} \), which is the concatenation of the decoder input \( \mathbf{s}_t^d \) and a set of task-specific features \( \mathbf{s}_t^s \). Specifically, we define
\[ 
\mathbf{s}_t^s = \{ \hat{p}_t^{\text{box}},\, \hat{\theta}_t^{\text{box}},\, \hat{p}_t^{\text{target}},\, \hat{\theta}_t^{\text{target}},\, \mathbf{d}_t \} \in \mathbb{R}^{19}, 
\] 
where \( \hat{p}_t^{\text{box}} \) and \( \hat{\theta}_t^{\text{box}} \) denote the distance and direction angle differences between the robot torso and the center of the box; \( \hat{p}_t^{\text{target}} \) and \( \hat{\theta}_t^{\text{target}} \) denote the distance and direction angle differences between the robot torso and the target position. Notably, the term \( \mathbf{d}_t \in \mathbb{R}^{15} \) comprises NSDF features, which are computed by a pre-trained network \( f_\theta \) evaluating the shortest distances between a selected target point and a set of mesh segments \cite{chen2025sampspatialanchorbasedmotion}, as shown in Fig. \ref{fig:4s}. In our implementation, we choose the center of the box as the target point $\mathbf{p}_t^{target}$ and the robot's upper links as the mesh collection, so that the nearest distances to the robot links are defined as: 
\begin{align}
\mathbf{d}_t = f_{\theta}(\mathbf{p}_t^{\text{target}}).
\end{align}

The task policy $\pi_{\mathrm{task}}$ operates in this latent space, producing a latent action $\mathbf{z}_t^{s}$ given the current task state $\mathbf{s}^{s}_t$:
\begin{align}
\mathbf{z}_t^{s} &= \pi_{\mathrm{task}}(\mathbf{s}^{s}_t), \\
\mathbf{a}_t^{s} &= \mathcal{D}(\mathbf{z}_t^{s} + \mu_t^p).
\end{align}

As shown in Fig. \ref{fig:2}(c), the learnable prior $\mathcal{R}$ generates a motion prior $\mu_t^p$. The output of $\pi_{\mathrm{task}}$ is then combined with this motion prior to form a new latent token, which is decoded into the final robot actions $\mathbf{a}_t^{s}$.

\subsubsection{Reward Design}

We design multiple reward functions to train the humanoid robot for the WBM task. The detailed descriptions of rewards on the smoothness and physical limitation are presented in Table \ref{tab:reward function}.

\textbf{Smoothness Reward}: The smoothness of the policy is encouraged by the reward term \( r_{\text{smooth}} = r_{\text{torque}} + r_{\text{acc}} + r_{\text{action}} \), where \( r_{\text{torque}} \) penalizes high joint torques, \( r_{\text{acc}} \) penalizes large actuator accelerations, and \( r_{\text{action}} \) penalizes rapid changes in action values. Collectively, these terms help guide the policy towards generating smooth and physically plausible actions.

\textbf{Physical Limitation Reward}: To protect the robot during deployment, we introduce \( r_{\text{limit}} = r_{\text{dof}} + r_{\text{slippage}} + r_{\text{feet}} \), where $r_{\text{dof}}$ encourages the joints to stay within their physical limits and discourages configurations near joint boundaries, $r_{\text{slippage}}$ enhances the stability of the robot. By limiting the contact force on the robot's foot, $r_{\text{feet}}$ prevents the robot from stepping too heavily on the ground, making its movements smoother.  

\textbf{Task Reward}: The total task reward is given by \( r_{\text{task}} = r_{\text{walk}} + r_{\text{carry}} + r_{\text{arm}} + r_{\text{NSDF}} \). 
The WBM task is decomposed into three distinct stages: approaching the object, embracing it, and transporting it to the target location. To support stage-specific reward allocation, a circular {\textit{pick-up zone}} of radius $d_p = 0.35$ m is centered around the object, acting as a transition boundary between approaching and manipulation. An annular region with inner radius $d_i = 3.5$ m and outer radius $d_o = 4.0$ m serves dual roles: it is used as the initial spawning area for the robot during the approaching phase and as the target delivery region for the object upon task completion.

In the first stage (when the robot approaches the object), the robot is not in the {\textit{pick-up zone}}. In this sense, $r_{\text{walk}}$ encourages the robot to move closer to the object, $r_{\text{carry}}$ and $r_\text{arm}$ are set to zero. The reward function is formulated as:
\begin{align}
r_{\text{walk}} = 
\begin{cases}
1,  \quad\quad \text{In pick-up zone},\vspace{0.5em}\\
\text{exp} (- \|\sigma\hat{p}_t^{\text{box}}\|_2 ) + \text{exp} (- \| \sigma\hat{\theta}_t^{\text{box}}\|_2) +\\
\quad\text{exp} (- \| \sigma \hat{v}_t^{\text{body}}\|_2), \,\, \text{Out of pick-up zone}.
\end{cases}
\end{align}
In the second and third stages when robot embraces and transports the object, it stays in the {\textit{pick-up zone}}, $r_{\text{carry}}$ encourages the box to move closer to the final target. In this case, the reward function is defined as:
\begin{align}
r_{\text{carry}} = 
\begin{cases}
0,  \quad\quad \text{Out of pick-up zone},\vspace{0.5em}\\ 
\text{exp} (- \|\sigma\hat{p}_t^{\text{target}}\|_2 ) + \text{exp} (- \| \sigma\hat{\theta}_t^{\text{target}}\|_2) +\\
\quad\text{exp} (- \| \sigma \hat{v}_t^{\text{box}}\|_2), \quad \text{In pick-up zone}.
\end{cases}
\end{align}
Additionally, $r_{\text{arm}}$ is used to adjust the arm position:
\begin{equation}
\begin{aligned}
r_{\text{arm}} = &\exp (-\sigma (\|\hat{p}_t^{\text{lh}}\|_2 + \|\hat{p}_t^{\text{rh}}\|_2) ) + \\
                 &\quad\exp (-\sigma (\|\hat{h}_t^{\text{lh}}\|_2 + \|\hat{h}_t^{\text{rh}}\|_2) ),
\end{aligned}
\end{equation}
where $\hat{p}_t^{\text{lh}}$ and $\hat{p}_t^{\text{rh}}$ are the distances from the left and right end-effectors to the object, respectively.

\begin{table}[tp]
\centering
\caption{Design of Reward Terms}
\begin{tabular}{ccc}
\toprule
\textbf{Reward Terms} & \textbf{Definition} & \textbf{Weight} \\
\midrule

Torque & $\left\|\mathbf{\tau}\right\|_2^2$ & -1e-7 \\[1ex]

Joint Acceleration & $\left\|\ddot{\mathbf{q}}\right\|_2^2$ & -2.5e-8 \\[1ex]

Action Rate & $\left\|\mathbf{a}_{t-1}-\mathbf{a}_t\right\|_2^2$ & -0.5 \\[1ex]

\midrule

Feet Slippage & $\sum \left| \mathbf{v}_{\text{foot}} \right| \cdot \left( \left| \mathbf{f}_{\text{contact}} \right| > 1 \right)$ & -0.05 \\[2ex]

Feet Contact Force & $\sum \left[ \max\left(0, \left|\mathbf{f}_{\text{contact}}\right| - F_{\text{max}}\right) \right]$ & -1e-5 \\[1ex]

Joint Angle Limitation & 
$\begin{aligned}
\sum \biggl[ & -\min\left(0, \mathbf{q} - \mathbf{q}_{\mathrm{lowerlimit}} \right) \\
& + \max\left(0, \mathbf{q} - \mathbf{q}_{\mathrm{upperlimit}} \right) \biggr]
\end{aligned}$ & -1e-3 \\[2ex] 

\bottomrule
\end{tabular}
\label{tab:reward function}
\end{table}

\subsubsection{Random Initialization}\label{sec:rand_init}
To address long-horizon tasks, we randomly generate initial scenarios that are categorized into three stages.
The first scenario type involves placing the target object on a table while randomly initializing the robot's position in the initial
spawning area. This configuration allows the training to focus primarily on the first stage of the task.
In the second scenario, the object remains placed on the table, but the robot's initial state is manually configured such that its upper joints are positioned in a pre-hugging posture, and the robot's position is initialized directly in the pick-up zone. This setup is designed to optimize training for the object pickup phase.
The third scenario initializes the robot's joints directly in a hugging posture, with the object generated in its arms. This configuration facilitates training for the final stage of the task.
During training, one of these three scenarios is randomly selected upon resetting the environment.

\section{Experiments}
In this section, we evaluate the proposed framework through both simulation and real-world experiments. The training environment is constructed using the Isaac Sim simulator, with 4096 parallel agents deployed for efficient data collection. The target object to be embraced is represented as a cylinder measuring $\Phi\,42\,\textup{cm} \times 40\,\textup{cm}$. Initially, the object is placed 4 meters away from the robot. The robot is then required to embrace the object and transport it to a target location situated 4 meters away from the object’s original position, thereby accomplishing an integrated whole-body manipulation task. Due to the absence of suitable open-source baselines in this domain, we mainly perform comparative experiments to validate the effectiveness and performance of the proposed modules. All training and experiments are conducted on a single workstation equipped with an NVIDIA RTX 4080 GPU.

\subsection{Validation of Our Framework}

To thoroughly evaluate the effectiveness of our proposed WBM framework, we design a set of experiments to analyze its core components. Specifically, we investigate two key aspects: the importance of the NSDF module in guiding the robot’s whole-body motion toward the target object, and the impact of randomization during early training on improving convergence efficiency.

\begin{figure}[tp]	
	\centering
        \includegraphics[width=0.96\linewidth]{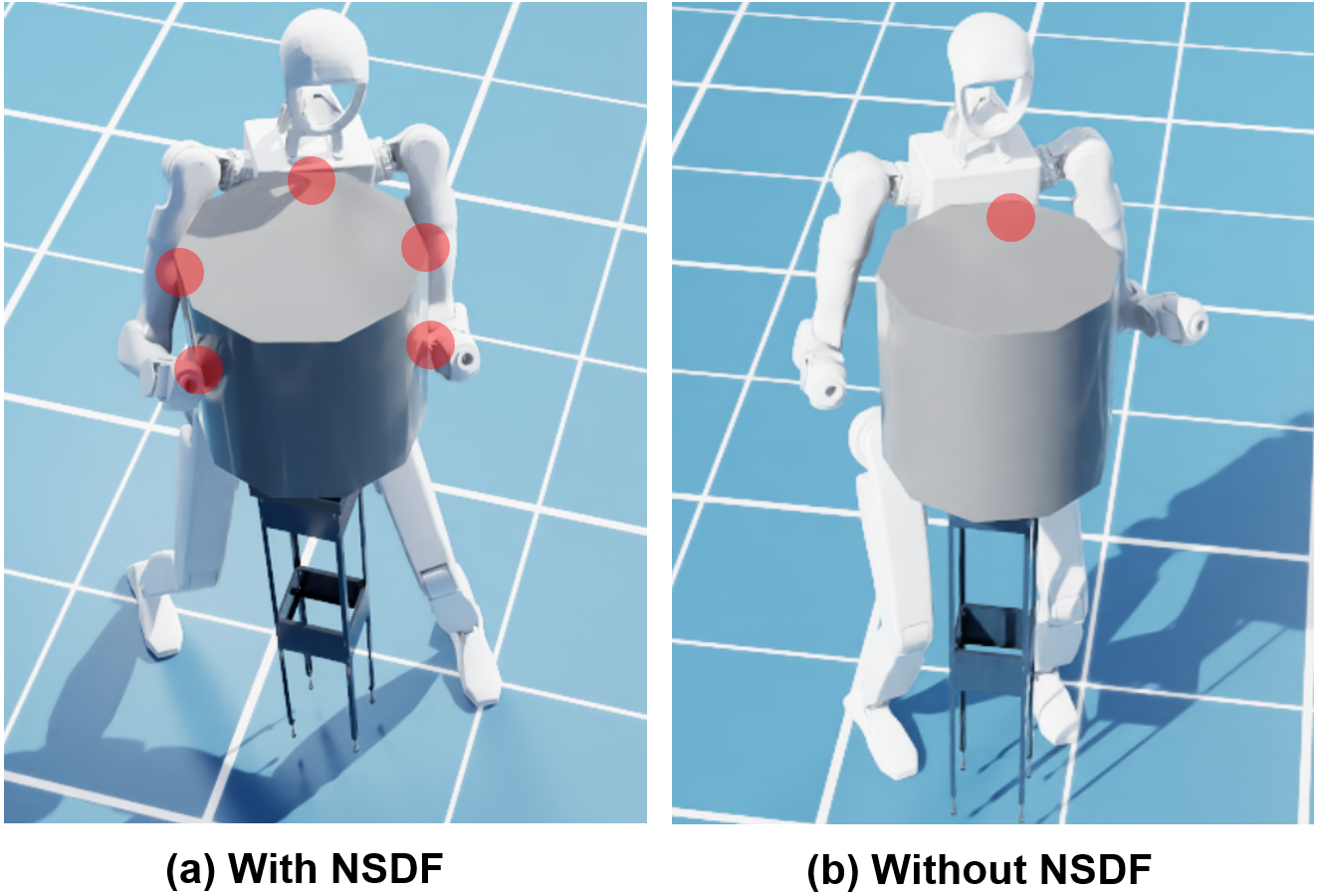}
	\caption{Effectiveness of the NSDF module. (a) Results with the NSDF module. (b) Results without the NSDF module. The red circles indicate contact regions. As shown, the model with NSDF generates more contact points during manipulation, significantly enhancing grasping stability and success rate.}
    \label{fig:NSDF_exp}
\end{figure}

\subsubsection{Effectiveness of the NSDF Module}

To evaluate the impact of this design, we adopt identical experimental settings without the NSDF-related observations from the input space and disable the NSDF reward term. As illustrated in Fig. \ref{fig:NSDF_exp}(a), after training for the same number of epochs, the robot trained with our proposed method successfully establishes multi-point contact using multiple upper-body joints during object transportation. This ability results in increased contact points with the object and significantly enhances manipulation stability, particularly when handling bulky items.
In contrast, Fig. \ref{fig:NSDF_exp}(b) depicts the trained policy without the NSDF module. During task execution, only the torso moves close to the object, while the arms fail to embrace properly. This leads to inferior policy performance and an inability to complete the transportation task.

\subsubsection{Evaluation of Randomized Initialization}

\begin{figure}[tp]	
	\centering
	\includegraphics[width=0.9\linewidth]
    {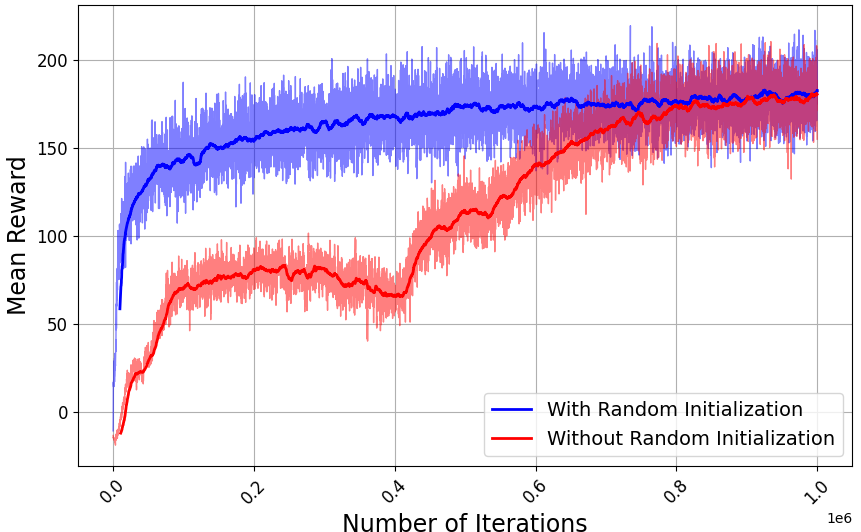}
	\caption{Comparison of mean rewards in the ablation study on multi-stage random initialization.
The red curve represents the policy with the proposed multi-stage random initialization, while the blue curve denotes the baseline without this mechanism. The policy with initialization (red) demonstrates smoother convergence and higher stability throughout training. In contrast, the baseline (blue) exhibits slower reward growth during the initial phase and greater instability.}
    \label{fig:random_exp}
\end{figure}

The proposed randomized initialization strategy mainly enhances state space exploration by resetting the robot to varied initial states at the beginning of each episode.
To evaluate its effectiveness, we perform the comparison against a baseline that starts each episode deterministically from a fixed initial state. As shown in Fig. \ref{fig:random_exp}, the proposed method (blue curve) demonstrates significantly improved convergence speed and stability compared to the fixed initialization baseline (red curve). Specifically, our method promotes balanced learning across all task phases, effectively overcoming the limitations of slow convergence and unstable reward progression observed in the baseline.

\begin{table}[tp]
    \centering
    \caption{Success Rate of Our Whole-Body Manipulation Policy for Objects with Different Sizes, Masses, and Shapes in MuJoCo.}
\begin{tabular}{ccccc}
    \toprule
     \textbf{Object}  &\makecell{\textbf{Size [cm$^3$]}} &  \makecell{\textbf{Mass} \\ \textbf{[kg]}}   & \makecell{\textbf{Success } \\ \textbf{rate [\%]}}  & \makecell{\textbf{NSDF}}\\
     \midrule
    Cylinder    & $\Phi\,42 \times 40$         &  3             & \textbf{0} & w\verb|/|o \\
    Cuboid      & $42\,\times 42 \times$ 42    &  3             & \textbf{0} & w\verb|/|o\\
    Sphere      & $\Phi\,42$                   &  3             & \textbf{0} & w\verb|/|o\\
    \midrule
    Cylinder    & $\Phi\,42 \times 40$         &  1             & 100 & w\\
    Cylinder    & $\Phi\,42 \times 40$         &  3             & 100 & w\\    
    Cylinder    & $\Phi\,42 \times 40$         &  7             & 93  & w\\  
    Cylinder    & $\Phi\,50 \times 40$         &  3             & 100 & w\\
    Cuboid      & $42\times 42 \times$ 42    &  1             & 100 & w\\
    Cuboid      & $42 \times 42 \times$ 42    &  3             & 100 & w\\
    Cuboid      & $42 \times 42 \times$ 42    &  7             & 80  & w\\  
    Cuboid      & $30 \times 30 \times$ 30    &  3             & 100 & w\\
    Sphere      & $\Phi\,42$                   &  1             & 100  & w\\
    Sphere      & $\Phi\,42$                   &  3             & 100  & w\\
    Sphere      & $\Phi\,42$                   &  7             & 87  & w\\
    Sphere      & $\Phi\,30$                   &  3             & 100  & w\\
     \bottomrule
\end{tabular}
    \label{tab:success_rate}
\end{table}

\begin{figure}[t]	
	\centering
	\includegraphics[width=1.0\linewidth]{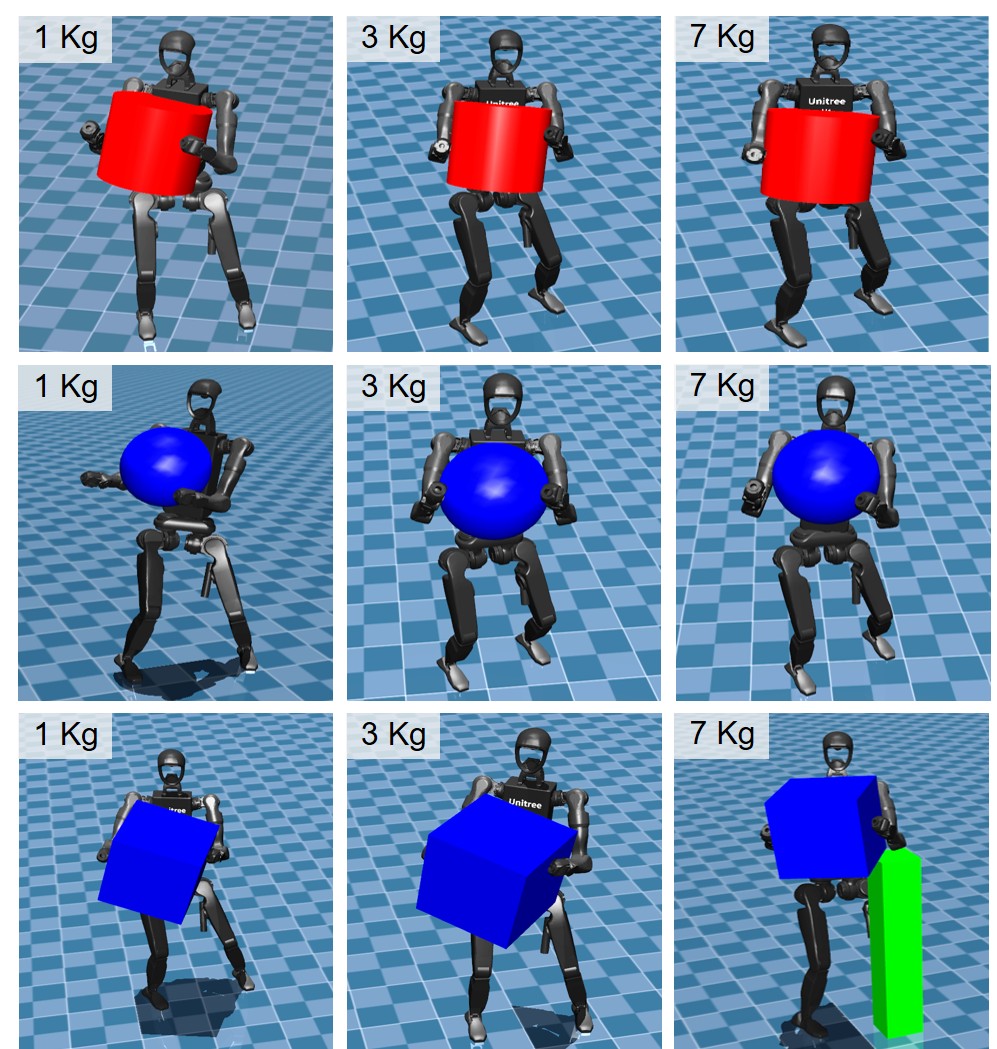}
\caption{Illustration of sim-to-sim experiments in Mujoco. The illustration compares the manipulation performance across three common object primitives: a cylinder, a cuboid, and a sphere. Objects are varied in both shape and weight to evaluate generalization. The red objects match the dimensions used during training, while those in blue represent unseen shapes for generalization testing. The green columns serve as pedestals for initial object placement. }
    \label{fig:mujoco}
\end{figure}

\begin{figure}[t]	
	\centering
    \includegraphics[width=1.0\linewidth]{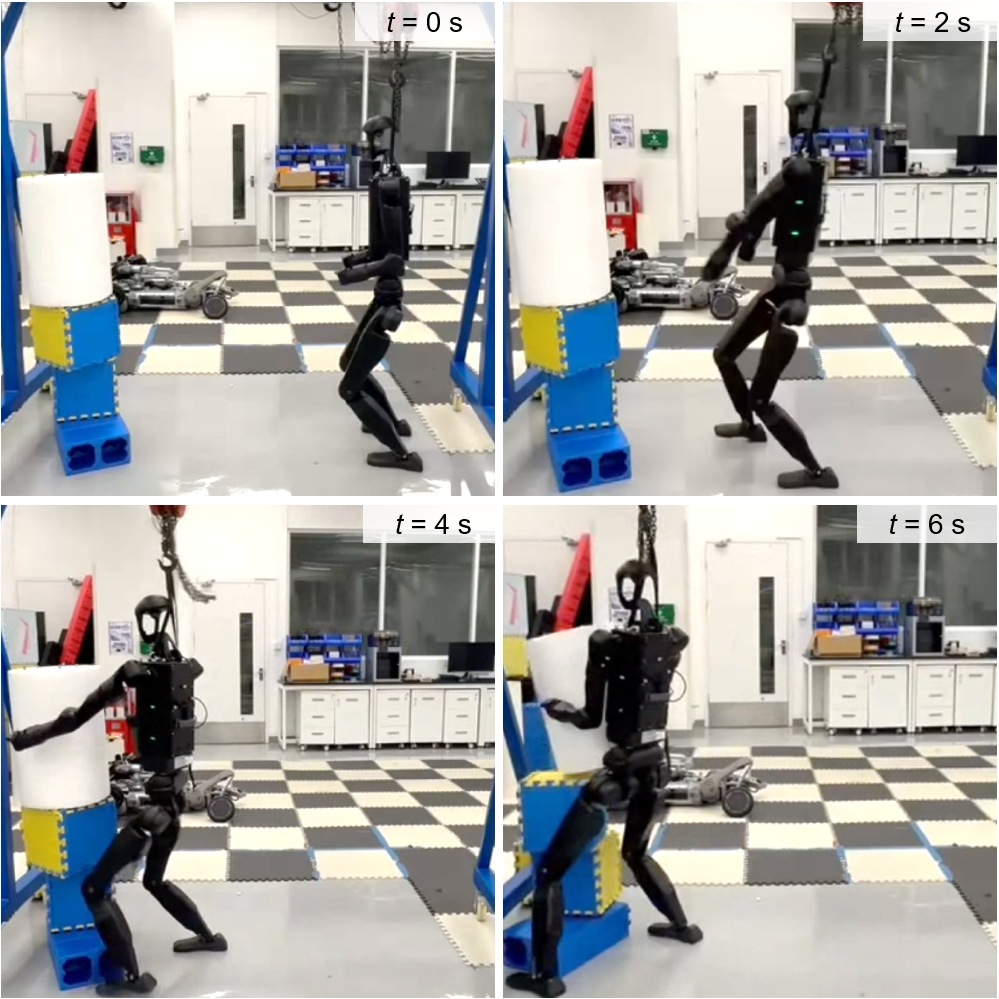}
\caption{Sim-to-real transfer of whole-body manipulation in H1-2 humanoid robot. The sequence illustrates the sim-to-real transfer of a whole-body manipulation task. From 0 to 2\,s, the H1 robot approaches the target object. At 4\,s, it positions itself to initiate contact and prepare for lifting. By 6\,s, the robot successfully lifts the object and begins locomotion to complete the task.}
    \label{fig:real}
\end{figure}

\subsection{Adaptability to Different Object Properties}

To comprehensively evaluate the robustness and generalization of the policy, we perform a sim-to-sim transfer assessment in MuJoCo, which enables standardized and reproducible testing across a wide range of physical scenarios, as illustrated in Fig. \ref{fig:mujoco}. In this environment, the policy is tested on objects with diverse sizes, masses, and geometries to evaluate its success rate in WBM tasks, even though it was trained solely on a fixed-size cylindrical object. For each trial, success is defined as the object’s center of mass remaining within 0.1\,m of the target location without falling. To ensure statistical reliability, the success rates are computed over 30 independent trials for each test condition.

The results summarized in Table \ref{tab:success_rate} reveal several key aspects of the policy's generalization capability. First, when the NSDF module is incorporated, the policy achieves and maintains a 100\% success rate across multiple object types, including cylinders, cuboids, and spheres, under standard mass (3\,kg) and size conditions. This demonstrates strong sim-to-sim transfer performance. In contrast, the policy exhibits complete failure (0\% success) across all shape categories without NSDF, highlighting the essential role of this module in capturing geometric and semantic attributes for stable manipulation. Furthermore, the policy shows notable robustness to mass variations, retaining high success rates even at 7\,kg, with 93\% for cylinders, 87\% for spheres, and 80\% for cuboids. The slightly lower performance observed for heavier cuboidal objects can be attributed to their flat surface morphology. Additionally, variations in object size, such as larger objects like the $\Phi$\,50\,cm cylinder and smaller ones like the $\Phi$\,30\,cm sphere, do not degrade performance. This further confirms the policy’s adaptability beyond the training distribution.

\subsection{Real-World Experiment}
In our real-world experiment, we deploy the trained policy on the Unitree H1-2 platform equipped with an onboard Intel i7 computer that executes all policy inference computations with the trained policy running at 50\,Hz, ensuring real-time control performance. The global poses of the robot and the embracing target are tracked in real time via a high-precision motion capture system. For the target object, we selected a cylinder with dimensions $\Phi\,40 \, \text{cm}\times60 \, \text{cm}$ to evaluate the policy's performance when handling objects of realistic size and structural properties.
As shown in Fig. \ref{fig:real}, the experiment showcases the robot's noteworthy capability in manipulation of a bulky object. The trial begins at 0\,s, with the robot initiating motion toward the object. It successfully enters the pick-up zone at 4\,s, then achieves a stable embracing of the object and moves into the transport phase by 6\,s. From the real-world experiments, they showcase the successful embracing performance toward bulky objects, while using the whole body of the humanoid robot including arms and torso. In this sense, the effectiveness of our proposed method is clearly demonstrated.

\section{Conclusions}
In this study, we propose an RL framework that integrates a pre-trained human motion prior and an NSDF, enabling humanoid robots to perform whole-body embracing tasks for bulky objects. By constructing a hierarchical control architecture, the method effectively coordinates the movements of robot arms and torso, to distribute contact forces across multiple contact points, which significantly enhances operational stability and load capacity. Experimental validation demonstrates that the proposed approach not only adapts to objects of various shapes, sizes, and weights in simulation but also transfers effectively to a real-world robot platform toward  WBM tasks with complex contacts.





\bibliographystyle{IEEEtran}
\bibliography{ref}

\end{document}